%% file: root.tex
\documentclass[letterpaper, 10 pt, conference]{ieeeconf}  

\IEEEoverridecommandlockouts                              

\title{\LARGE \bf
Physical Diffusion Policy via MPM-Guided 3D Gaussian Simulation
}

\author{Can Wang$^{1}$, Qiyuan Qiao$^{1}$, Lei Liu$^{1}$, Ge Yuan$^{1}$, Xinhui Liu$^{1}$, and Dong Xu$^{1,*}$%
\thanks{*Dong Xu is the corresponding author}%
\thanks{$^{1}$The University of Hong Kong, Hong Kong SAR, China.}%
}
\usepackage{algorithm}
\usepackage{algorithmic}
\usepackage{float}
\usepackage{multirow}
\usepackage{amssymb} 
\usepackage{booktabs}
\usepackage{graphicx}
\usepackage{threeparttable}

\usepackage{pifont}

\newcommand{\cmark}{\ding{51}}
\newcommand{\xmark}{\ding{55}}

\usepackage{makecell}

\usepackage{bibentry}
\usepackage{amsmath}
\usepackage{xspace}

\begin{document}

\maketitle
\thispagestyle{empty}
\pagestyle{empty}

\input{sections/0_abstract}

\input{sections/1_intro}
\input{sections/2_related}

\input{sections/3_preliminary}

\input{sections/3_method}
\input{sections/4_exp}
\input{sections/5_con}

\addtolength{\textheight}{-12cm}   








\bibliographystyle{IEEEtran}
\bibliography{tormbib}

\end{document}

%% file: sections/0_abstract.tex
\begin{abstract}

Learning robust manipulation policies for diverse everyday objects remains challenging because objects with different materials exhibit distinct deformation and contact dynamics. While diffusion policies provide a powerful framework for modeling multimodal manipulation behaviors, they typically lack explicit physical reasoning and may produce averaged actions when facing heterogeneous objects. We present a physics-aware diffusion policy that augments visual imitation learning with object physical understanding. Our framework recovers material properties from visual demonstrations by coupling 3D Gaussian Splatting with differentiable Material Point Method simulation, where video guidance and particle-level constraints enable accurate physical parameter estimation. The estimated physics conditions are further integrated into a material-aware Mixture-of-Experts diffusion policy, allowing adaptive behavior generation across different physical regimes. For practical deployment, we introduce a physics memory bank that provides efficient physical retrieval without requiring online simulation optimization. Experiments on a wide range of simulated and real-world objects demonstrate improved manipulation success and generalization over existing diffusion-based approaches. These results show that incorporating physical knowledge is critical for scalable robot manipulation in diverse environments.

\end{abstract}

%% file: sections/1_intro.tex
\section{Introduction}

Robot manipulation in unstructured environments requires the ability to interact with objects that exhibit diverse physical properties. While humans can effortlessly adapt their behaviors by reasoning about material characteristics such as stiffness, deformability, and compliance, robots typically rely on visual observations and previously collected demonstrations to infer appropriate actions. This gap becomes particularly challenging when manipulating everyday objects, where even visually similar objects may require fundamentally different interaction strategies due to variations in their underlying physical dynamics.

Diffusion policies~\cite{chi2025diffusion} have recently emerged as a powerful framework for robot manipulation by learning expressive action distributions from human demonstrations. Recent efforts have further improved diffusion policies by incorporating richer geometric representations, such as 3D point clouds and multi-view spatial features, to enhance scene understanding and manipulation precision~\cite{liu2026twin,lv2026kinematics,wang2024gendp,ze2025generalizable,cao2025mamba,yang2024equibot,ke20243d,ze20243d}. Nevertheless, existing diffusion policies primarily treat objects as visual entities and lack explicit awareness of their physical properties. This limitation becomes particularly pronounced in real-world manipulation, where objects with different materials can exhibit substantially different deformation and contact behaviors. Without physical understanding, a single policy may average incompatible interaction patterns, resulting in suboptimal behaviors when encountering heterogeneous objects.

A natural solution is to provide manipulation policies with explicit physical knowledge of objects. However, obtaining accurate physical properties from visual observations remains challenging. Traditional system identification approaches typically require carefully designed interactions, force measurements, or object-specific simulators, which are difficult to scale to diverse real-world objects~\cite{calandra2018more,zheng2025survey,chen2025learning}. Recent advances in differentiable simulation~\cite{hu2019chainqueen,chen2022diffsrl,si2024difftactile,xie2024physgaussian,zhao2025physsplat,lin2025omniphysgs} provide an opportunity to infer physical parameters by optimizing simulation outcomes against observations. 
Nevertheless, existing methods are primarily designed for physical parameter estimation or sim-to-real transfer, rather than enabling robots to explicitly reason about object physics during policy learning. As a result, the challenge of incorporating physical understanding into general-purpose robot manipulation remains largely unresolved.

In this work, we propose a physics-aware manipulation framework that bridges visual observations, differentiable physical simulation, and diffusion-based control. Given multi-view observations of an object, we construct a unified 3D Gaussian Splatting (3DGS)~\cite{kerbl20233d} and Material Point Method (MPM)~\cite{sulsky1994particle,sulsky1995application,sulsky1996axisymmetric} representation, enabling differentiable simulation and rendering. By optimizing the simulated dynamics against demonstration videos with particle-level physical constraints, our framework estimates object-specific physical parameters, including elasticity, plasticity, and viscosity-related properties. These physical properties are then used as conditions for a physics-aware diffusion policy. To better handle diverse material behaviors, we introduce a material-aware Mixture-of-Experts (MoE) architecture, where a learned router dynamically selects material-specific experts to generate physics-consistent manipulation trajectories. Furthermore, we develop a physics memory bank that stores previously estimated physical properties, allowing efficient deployment without expensive online simulation optimization.


Our contributions are summarized as follows:
\begin{itemize}
    \item We introduce a physics-aware manipulation framework that integrates differentiable physical simulation with diffusion policies, enabling robot manipulation conditioned on estimated object physical properties.
    \item We propose a differentiable 3DGS-MPM representation with video-guided optimization and particle-level constraints to recover material properties from visual observations.
    \item We design a material-aware MoE diffusion policy that adaptively selects specialized experts for different physical regimes, improving manipulation generalization across heterogeneous objects.
    \item We construct a physics memory bank for efficient deployment, avoiding costly online physical optimization while preserving physics-consistent manipulation performance.
\end{itemize}

%% file: sections/2_related.tex
\section{Related Work}

\noindent \textbf{Diffusion Models in Robotics.}
Diffusion Policy~\cite{chi2025diffusion} establishes diffusion models as an effective framework for robot manipulation.
Building upon this framework, subsequent works have explored different directions to improve policy representation and learning effectiveness. For example, DPPO~\cite{ren2025diffusion} further bridges diffusion policies with reinforcement learning, enabling effective policy improvement beyond imitation learning. 
More recently, latent-space reinforcement learning has also been introduced to steer pretrained diffusion policies toward downstream objectives while preserving their generative capabilities~\cite{wagenmaker2025steering}.
Another active research direction focuses on incorporating richer 3D scene representations into diffusion policies. DP3~\cite{ze20243d} uses point-cloud representations to improve geometric perception during manipulation. 3D Diffuser Actor~\cite{ke20243d} directly conditions policy generation on 3D visual representations, while EquiBot~\cite{yang2024equibot} exploits SE(3)-equivariant representations to improve spatial generalization. 
More recent works further investigate efficient sequence modeling for 3D diffusion policies using hybrid state-space architectures~\cite{cao2025mamba} and extend DP3 to generalizable humanoid manipulation through egocentric 3D representations~\cite{ze2025generalizable}. 
Despite their strong geometric reasoning ability, existing diffusion policies largely ignore object physical properties. 
In contrast, our method explicitly conditions diffusion policies on estimated object physics, enabling material-aware manipulation.

\paragraph{Simulation in Robot Learning.}
Simulation has long served as a fundamental tool for robot learning by enabling scalable data collection, policy optimization, and safe exploration before real-world deployment. 
Existing sim-to-real methods primarily focus on reducing the domain gap between simulated and real environments through techniques such as domain randomization, system identification, and adaptive policy learning~\cite{peng2018sim,james2019sim,memmel2024asid,yu2019sim,zhang2025dynamics,scheikl2022sim,ju2022transferring}. More recent simulation platforms further improve the realism of contact dynamics and large-scale parallel training, allowing manipulation policies to be trained efficiently in simulation and transferred to physical robots~\cite{makoviychuk2021isaac,nvidia2021isaacsim}.
In contrast, our work leverages differentiable simulation from a different perspective. Rather than optimizing policies inside the simulation or improving simulator realism, we employ simulation as a physical reasoning module to estimate object-specific material parameters from visual demonstrations. The recovered physical properties are subsequently injected into a diffusion policy as physics conditions, enabling adaptive manipulation across diverse material categories without relying on online simulation during deployment.

\paragraph{Physics-based 3D Gaussian Simulation.} 
The emergence of 3DGS has enabled recent efforts to integrate explicit 3D scene representations with physics-based simulation for modeling dynamic object behaviors. Several works couple 3DGS with MPM to achieve physically plausible deformation, differentiable rendering, and editable dynamic scenes~\cite{xie2024physgaussian,zhao2025physsplat,lin2025omniphysgs}. 
By attaching Gaussian primitives to simulated particles, these methods preserve photorealistic rendering quality while enabling realistic physical animation under predefined material properties.
Despite their impressive visual fidelity, existing 3DGS-based physics simulation methods primarily target computer graphics applications such as dynamic scene reconstruction, physical animation, and interactive editing. They generally assume that object material parameters are known a priori and focus on the forward simulation of object dynamics under prescribed constitutive models. Consequently, these approaches are not designed for robotic manipulation. In contrast, our method formulates the problem from an inverse perspective. Instead of using predefined physical parameters to animate a reconstructed scene, we leverage differentiable MPM-3DGS simulation to recover object-specific material parameters from visual demonstrations. The estimated physical properties are subsequently used to condition a diffusion policy, bridging differentiable physical simulation with physics-aware robot manipulation.

%% file: sections/3_preliminary.tex
\section{Preliminary}

\noindent \textbf{Feed-forward 3D Gaussian Splatting:}
Feed-forward 3D Gaussian Splatting (FF-3DGS) \cite{Wang_2026_CVPR,charatan2024pixelsplat} aims to reconstruct an explicit Gaussian scene representation directly from sparse input images, which circumvents the time-consuming per-scene optimization inherent in conventional 3DGS \cite{kerbl20233d}. Formally, given a set of multi-view images $\mathcal{I}=\{I_i\}_{i=1}^{N}$, FF-3DGS leverages a neural network $\mathcal{F}$ to predict a collection of 3D Gaussian primitives via a single forward pass:
\begin{equation}
\mathcal{G}=\{G_k\}_{k=1}^{K}
=
\mathcal{F}(\mathcal{I}),
\end{equation}
where each Gaussian $G_k$ is parameterized by its center $\boldsymbol{\mu}_k$, opacity $\alpha_k$, rotation $\mathbf{r}_k$, scale $\mathbf{s}_k$, and color feature $\mathbf{c}_k$. 
By directly regressing geometric and radiometric attributes from visual observations, FF-3DGS offers superior computational efficiency and cross-scene generalization compared to its optimization-based counterparts. In this work, we adopt FF-3DGS to instantiate an object-centric Gaussian representation from multi-view inputs, yielding a high-fidelity rendering scaffold that is naturally amenable to downstream coupling with MPM-based differentiable physical simulation.
\\

\noindent \textbf{Material Point Method:} The Material Point Method (MPM)~\cite{sulsky1994particle,sulsky1995application,sulsky1996axisymmetric} is a hybrid Eulerian–Lagrangian method for simulating continuum dynamics. It represents material bodies as a set of Lagrangian particles while using an Eulerian background grid to compute momentum transfer and spatial derivatives. Each particle carries physical states such as position $\mathbf{x}_p$, velocity $\mathbf{v}_p$, mass $m_p$, volume $V_p$, and deformation gradient $\mathbf{F}_p$. In each time step, particle states are transferred to grid nodes for force evaluation (Particles-to-Grid, P2G), and the updated grid velocities are subsequently interpolated back to update the particles (Grid-to-Particles, G2P). The kinematic update of the deformation gradient is given by
\begin{equation}
\mathbf{F}_p^{t+1}
=
\left(\mathbf{I}+\Delta t \nabla \mathbf{v}_p^{t+1}\right)
\mathbf{F}_p^{t}.
\end{equation}
Due to its robustness in handling large deformations and topological transitions, MPM is exceptionally effective for simulating deformable objects. Crucially, we bind object-centric 3D Gaussian primitives directly onto MPM particles. This integration yields a fully differentiable simulation pipeline, empowering our framework to accurately estimate physical parameters and provide physically consistent regularization for robotic manipulation.

%% file: sections/3_method.tex
\section{Method}

\begin{figure*}
    \centering
    \includegraphics[width=\linewidth]{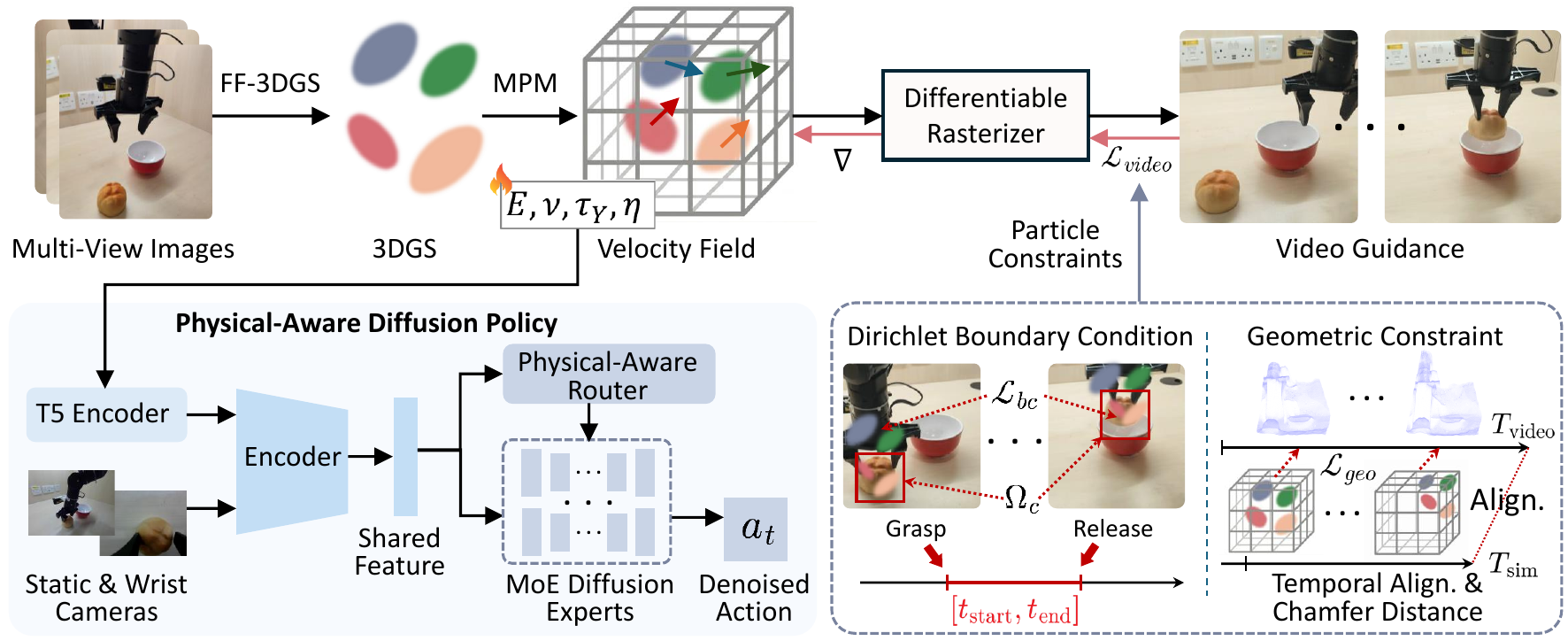}
    \caption{
    \textbf{Overview of the proposed training framework.}
    Given multi-view images, we first estimate the physical properties of the manipulated object by combining feed-forward 3D Gaussian Splatting (FF-3DGS) with MPM via a differentiable physical simulation.
    The simulator is coupled with a differentiable rasterizer and optimized under video guidance, where gradients are back-propagated through the simulation. Physics-based constraints, including Dirichlet boundary conditions for manipulation events and geometric consistency via temporal Chamfer distance, further regularize the estimated physical parameters by aligning simulated particle dynamics with demonstration videos.
    The learned physical parameters are then injected into a physical-aware diffusion policy as physics conditions, together with language instructions and visual observations, enabling the policy to predict physics-consistent robot actions and gripper states.
    }
    \label{fig:framework}
\end{figure*}

We propose a physics-aware manipulation framework, the training pipeline of which is illustrated in Fig.~\ref{fig:framework}. Given multi-view images, we first construct an MPM-guided 3DGS representation for differentiable physical simulation and estimate object physical parameters through video-guided optimization with a differentiable rasterizer. During optimization, we impose particle-level constraints, including Dirichlet boundary conditions and geometric consistency constraints, to improve the stability and accuracy of the simulation. We then train a physics-aware diffusion policy that takes language instructions, static and wrist-camera observations, together with the estimated physical parameters, as input to predict denoised robot actions and gripper states. 
Finally, to facilitate efficient inference, we construct a memory bank that archives a diverse set of object physical properties. During deployment, this memory bank is queried to retrieve the most relevant physical parameters, enabling the diffusion policy to generate physics-consistent manipulation actions while bypassing the need for computationally expensive physics optimization.

\subsection{Differentiable Physical Scene Representation}
\label{sec:scene_representation}

Our goal is to build a scene representation that supports both high-fidelity rendering and differentiable physical simulation. We capture a hemispherical video of the robotic scene and uniformly sample 64 frames. A ZFormer FF-3DGS~\cite{Wang_2026_CVPR} is then used to reconstruct a Gaussian representation $\mathcal{G}=\{G_k\}_{k=1}^{K}$.
To enable physical simulation, we instantiate an MPM particle system $\mathcal{P}=\{P_k\}_{k=1}^{K}$, where each particle maintains state variables including velocity $\mathbf{v}_k$, mass $m_k$, volume $V_k$, and deformation gradient $\mathbf{F}_k$, along with constitutive parameters $\boldsymbol{\theta}=\{E,\nu,\tau_Y,\eta\}$, where $E$ denotes Young's modulus, $\nu$ is Poisson's ratio, $\tau_Y$ is the yield stress, and $\eta$ is the viscosity. The system is evolved using a differentiable MPM solver~\cite{xie2024physgaussian}.
We then establish a one-to-one correspondence between Gaussians and particles. Each Gaussian center follows particle motion:
\begin{equation}
\boldsymbol{\mu}_k^t = \mathbf{x}_k^t,
\end{equation}
The Gaussian covariance is updated according to the particle deformation via polar decomposition $\mathbf{F}_k^t=\mathbf{R}_k^t\mathbf{S}_k^t$:
\begin{equation}
\mathbf{\Sigma}_k^t = \mathbf{R}_k^t \mathbf{\Sigma}_k^0 (\mathbf{R}_k^t)^\top.
\end{equation}
This coupling ensures consistent evolution between physical dynamics and Gaussian-based rendering, enabling differentiable rendering at each simulation step.

Given material parameters $\boldsymbol{\theta}$ and external manipulation actions, the MPM simulator predicts particle trajectories,
\begin{equation}
\{\mathbf{x}_k^t,\mathbf{v}_k^t,\mathbf{F}_k^t\}_{t=1}^{T}
=
\mathrm{MPM}(\mathcal{P},\boldsymbol{\theta}),
\end{equation}
which are directly propagated to the Gaussian primitives for differentiable rendering, where $\mathbf{v}_k^t$ is the velocity.
Since both the simulator and rasterizer are fully differentiable, gradients from image reconstruction losses can be back-propagated through rendered Gaussians to particle dynamics and ultimately to the unknown physical parameters.

\subsection{Physics-Guided Differentiable Simulation}
\label{sec:differentiable_simulation}

Based on the differentiable physical scene representation, we recover the underlying physical parameters
$\boldsymbol{\theta}$ by minimizing the discrepancy between the simulated dynamics and the observed motion under both visual and physical constraints.
To improve the convergence of the optimization, we first initialize
$\boldsymbol{\theta}$ using a Vision-Language Model (VLM)~\cite{bai2025qwen3}, which provides a coarse estimate of the object's material properties. Since the predictions of the VLM may be inaccurate or physically implausible, and the subsequent optimization steps may exceed valid physical boundaries, we introduce a material-aware parameter clipping strategy. Specifically, for each physical parameter $\theta_i$, we define a feasible interval based on the inferred material category. We then project the parameter back into this valid range both immediately after the initial VLM estimation and after each subsequent gradient update:
\begin{equation}
\theta_i \leftarrow
\operatorname{clip}
\left(
\theta_i,\,
\theta_i^{\mathrm{low}},\,
\theta_i^{\mathrm{up}}
\right),
\label{eq:param_clip}
\end{equation}
where $\theta_i^{\mathrm{low}}$ and $\theta_i^{\mathrm{up}}$ denote the lower and upper bounds of the parameter, respectively. These bounds are determined by the material type (e.g., elastic, metal, or plasticine) and restrict the optimization to physically meaningful regions, preventing unrealistic parameter updates and improving the stability of differentiable simulation.

\paragraph{Temporal Alignment.}
Before enforcing video  supervision and geometric consistency, we temporally align the simulation trajectory with the demonstration video to avoid de-synchronization caused by the discretized simulation step size. Let $\Delta t$ denote the temporal interval between adjacent video frames, $T_{\mathrm{video}}$ the video duration, and $N_{\mathrm{kf}}$ the number of keyframes used in each optimization stage. We define $M_{\min}$ and $M_{\max}$ as the minimum and maximum allowable numbers of simulation stages, respectively. The number of simulation stages is determined as
\begin{equation}
M(\Delta t,T_{\mathrm{video}})
=
\operatorname{clip}
\left(
\left\lfloor
\frac{T_{\mathrm{video}}}
{N_{\mathrm{kf}}\cdot\Delta t}
\right\rfloor,
M_{\min},
M_{\max}
\right),
\end{equation}
such that the final simulation time $T_{\mathrm{sim}}$ is aligned with the video duration. 

\paragraph{Video-Guided Rendering Loss.}

To recover physical properties, we record a monocular video of a robotic arm performing the target manipulation from a fixed viewpoint and use it as guidance. Since the temporal correspondence between the simulated and observed frames is unknown, we compare the rendered trajectory with the recorded video using a video guidance model that evaluates temporal consistency and visual similarity:
\begin{equation}
\mathcal{L}_{video}
=
\mathcal{D}_{video}
(\{\mathbf{I}_{sim}^{t}\}_{t=1}^{T},
\{\mathbf{I}_{ref}^{t}\}_{t=1}^{T}).
\end{equation}
Here, $\{\mathbf{I}_{ref}^{t}\}_{t=1}^{T}$ denotes the monocular guidance video recorded during the real-robot manipulation.

\paragraph{Geometric Consistency Constraint.}
Based on the above temporal correspondence, we enforce geometric consistency between the simulated particles and the observed object geometries:
\begin{equation}
\mathcal{L}_{geo}
=
\sum_{t}
\mathcal{D}_{\mathrm{Chamfer}}
(
\mathcal{X}_{sim}^{t},
\mathcal{X}_{obs}^{t}
),
\end{equation}
where $\mathcal{X}_{sim}^{t}$ and $\mathcal{X}_{obs}^{t}$ denote the temporally aligned simulated and observed object geometries, respectively, with $\mathcal{X}_{obs}^{t}$ being extracted from DA3~\cite{ICLR2026_e4cd5012}.

\paragraph{Manipulation Boundary Constraint.}
Although video supervision provides visual guidance, it does not explicitly enforce the physical interaction between the gripper and the object. Therefore, we introduce a Dirichlet boundary constraint on particles influenced by the robot gripper. Following the demonstrated grasping motion, particles inside the contact region are constrained to follow the observed gripper trajectory during the grasping phase $[t_{start}, t_{end}]$, and the constraint is removed after release:
\begin{equation}
\mathcal{L}_{bc}
=
\sum_{t=t_{start}}^{t_{end}}
\sum_{k\in\Omega_c}
\|
\mathbf{x}_{k}^{t}
-
\hat{\mathbf{x}}_{k}^{t}
\|_2^2 ,
\end{equation}
where $\Omega_c$ denotes the constrained particle set and $\hat{\mathbf{x}}_{k}^{t}$ represents the desired particle position induced by the gripper motion.

The final optimization objective combines visual supervision and particle-level physical constraints:
\begin{equation}
\mathcal{L}
=
\mathcal{L}_{video}
+
\lambda_{bc}\mathcal{L}_{bc}
+
\lambda_{geo}\mathcal{L}_{geo}.
\end{equation}
Since the MPM simulator and Gaussian rasterizer are differentiable, gradients propagate through rendering and dynamics to optimize $\boldsymbol{\theta}$ for end-to-end estimation of material properties and manipulation dynamics.

\subsection{Physical-Aware Diffusion Policy}
\label{sec:diffusion_policy}

Objects with different material properties exhibit substantially different deformation and contact behaviors, which often correspond to distinct manipulation strategies. A single diffusion policy trained across heterogeneous materials may average these action distributions and produce sub-optimal trajectories. Thus, we introduce a material-aware mixture-of-experts (MoE) diffusion policy with four experts corresponding to elastic, foam, plastic, and rigid materials. The estimated physical parameters and visual observations are encoded into a shared condition feature, which is used by a physical-aware
router $G$ to predict the routing weights.

Let $\mathcal{M}=\{\mathrm{ela},\mathrm{foam},\mathrm{pla},\mathrm{met}\}$ denote the set of material experts. The routing weights are computed as
\begin{equation}
w_m
=
\operatorname{Softmax}
\left(
G(\mathbf{c}_t)
\right), \sum_{m\in\mathcal{M}} w_m=1
\end{equation}
where $\mathbf{c}_t$ is the shared condition feature.
Following the standard diffusion formulation, a clean action sequence $\mathbf{a}_0$ is perturbed at diffusion timestep $k$ as
\begin{equation}
\mathbf{a}_k
=
\sqrt{\bar{\alpha}_k}\mathbf{a}_0
+
\sqrt{1-\bar{\alpha}_k}\boldsymbol{\epsilon},
\end{equation}
where $\boldsymbol{\epsilon}\sim\mathcal{N}(\mathbf{0},\mathbf{I})$.
Each material expert predicts the noise independently,
\begin{equation}
\hat{\boldsymbol{\epsilon}}_m
=
\epsilon_{\phi_m}
\left(
\mathbf{a}_k,
k,
\mathbf{c}_t
\right),
\qquad
m\in\mathcal{M}.
\end{equation}
The final noise prediction is obtained by aggregating the outputs of all experts according to the routing weights,
\begin{equation}
\hat{\boldsymbol{\epsilon}}
=
\sum_{m\in\mathcal{M}}
w_m
\hat{\boldsymbol{\epsilon}}_m.
\end{equation}
The policy is optimized using the denoising objective
\begin{equation}
\mathcal{L}_{diff}
=
\mathbb{E}_{\mathbf{a}_0,k,\boldsymbol{\epsilon}}
\left[
\left\|
\boldsymbol{\epsilon}
-
\hat{\boldsymbol{\epsilon}}
\right\|_2^2
\right].
\end{equation}
By using material-specific diffusion experts, the policy reduces action averaging and generates trajectories that are better aligned with object physics.

\subsection{Physics-Conditioned Online Inference}
\label{sec:online_inference}

Optimizing physical parameters with differentiable simulation is computationally expensive and thus impractical during deployment. We therefore construct a physics memory bank that associates multi-view visual features with the physical parameters estimated offline.

For each object $i$, we extract DINOv2~\cite{oquab2023dinov2} features from the static and wrist-camera observations and store them with the optimized physical parameters:
\begin{equation}
\mathcal{B}
=
\left\{
\left(
\mathbf{f}_{i}^{s},
\mathbf{f}_{i}^{w},
\boldsymbol{\theta}_{i}
\right)
\right\}_{i=1}^{N_{\mathcal{B}}},
\end{equation}
where $\mathbf{f}_{i}^{s}
=
\Phi\left(\mathbf{I}_{i}^{s}\right)$, $\mathbf{f}_{i}^{w}
=
\Phi\left(\mathbf{I}_{i}^{w}\right),$ and $\Phi(\cdot)$ denotes the frozen DINOv2 encoder.

At test time, we extract query features $\mathbf{f}_{q}^{s}$ and $\mathbf{f}_{q}^{w}$ from the initial static and wrist-camera observations. The similarity between the query object and the $i$-th memory entry is computed by averaging the cosine similarities:
\begin{equation}
s_{q,i}
=
0.6
\frac{
\left(\mathbf{f}_{q}^{s}\right)^{\top}\mathbf{f}_{i}^{s}
}{
\left\|\mathbf{f}_{q}^{s}\right\|_{2}
\left\|\mathbf{f}_{i}^{s}\right\|_{2}
}
+
0.4
\frac{
\left(\mathbf{f}_{q}^{w}\right)^{\top}\mathbf{f}_{i}^{w}
}{
\left\|\mathbf{f}_{q}^{w}\right\|_{2}
\left\|\mathbf{f}_{i}^{w}\right\|_{2}
}.
\end{equation}
Following the relative retrieval performance of the two views (Table~\ref{tab:retrieval_performance}), we assign a larger weight to the static view and use fixed fusion weights of 0.6 and 0.4 for the static and wrist features, respectively.
We retrieve the nearest memory entry and use its physical parameters as the estimate for the target object.
The retrieved parameters are then provided to the physics-aware diffusion policy. This design avoids online physical optimization and enables efficient physics-conditioned manipulation.

%% file: sections/4_exp.tex
\section{Experiments}

\paragraph{Experimental Setup.} 
We evaluate in both simulation and real-world environments. In simulation, we build the experiments in an Isaac Lab environment~\cite{makoviychuk2021isaac,nvidia2021isaacsim} containing diverse objects with different physical properties, including ceramic cups, metal cubes, plastic cubes, rubber balls, sponges, teddy bear dolls, and wooden objects. 
For real-world evaluation, we collect experiments on daily objects, including hard candies, circular and elongated bread, stuffed toys, foam balls, washing sponge, ping-pong balls, tennis balls, tissue packages, tape rolls, toothpaste tubes, bananas, grapes, springs, shuttlecocks, wooden blocks, and disposable cups. 
These objects cover a wide range of material categories, including elastic, foam, plasticine, and rigid or near-rigid materials.

We evaluate two manipulation tasks: pick-and-place and stacking. For the pick-and-place task, all objects are used for evaluation. For the stacking task, we select a subset of objects requiring more precise interaction, including Shuttlecocks, tissue packages, disposable cups, washing sponges and wooden cubes in the real-world setting. 

For data collection, we collect 300 demonstrations in simulation using a scripted state-machine controller. In the real-world environment, we manually collect 100 demonstrations using a PS5 controller for teleoperation. All demonstrations are recorded at 20 Hz and contain synchronized visual observations, robot states, and action trajectories.

\paragraph{Implementation Details.} We integrate 3DGS with a GPU-accelerated MPM simulator~\cite{xie2024physgaussian}. The MPM domain is discretized with a uniform $100^3$ grid. Particle velocities are initialized to zero, and deformation gradients are initialized as identity matrices. We optimize the physical parameters for 30 epochs using 8 key frames per sequence. The simulation timestep is set to 0.04 s, and input video frames are processed at 25 FPS.
Instead of training object-specific policies, we train task-level generalist policies, with one model for pick-and-place and another for stacking. Our physical-aware diffusion policy is built upon an image-conditioned diffusion policy with DDIM sampling. The policy takes observations from a fixed table camera and a wrist-mounted camera, together with a 7-dimensional robot state vector containing end-effector pose and gripper state. The action space consists of an 8-dimensional end-effector goal pose and gripper command.

We train the material-aware MoE diffusion policy in two stages. In the first stage, we independently train four expert diffusion policies corresponding to elastic, foam, plastic, and rigid materials using demonstrations from their respective material categories. In the second stage, we initialize the MoE policy with the pretrained expert weights, introduce the physics-aware router, and jointly fine-tune the router and all expert networks end-to-end. Both stages use the same optimization settings. Specifically, all models are trained for 200 epochs with a batch size of 32 and a learning rate of $1\times10^{-4}$ on a single NVIDIA A100 GPU. 

We compare our method against the native diffusion policy (DP)~\cite{chi2025diffusion} and DP3 ~\cite{ze20243d}, which incorporate 3D point information to improve performance. For a fair comparison, all methods are trained using the same demonstration datasets and adopt the same Diffusion Policy backbone. We use the task success rate as the primary evaluation metric. In simulation, we report the success rate over 120 trials per task in Isaac Lab. For real-world experiments, we conduct 30 trials for each task.

\subsection{Physical Parameter Estimation}
\label{sec:exp_param_estimation}

\begin{table}[t]
    \centering
    \caption{
    Quantitative evaluation of physical parameter estimation.
    We report log-MAE for Young's modulus $E$, yield stress $\tau_Y$, and viscosity $\eta$, and MAE for Poisson's ratio $\nu$.
    Lower values indicate better estimation accuracy.
    }
    \label{tab:physical_parameter_estimation}
    \resizebox{\columnwidth}{!}{
    \begin{tabular}{lccccc}
        \toprule
        Method
        & $E \downarrow$
        & $\nu \downarrow$
        & $\tau_Y \downarrow$
        & $\eta \downarrow$
        & Avg. $\downarrow$ \\
        \midrule
        w/o VLM Init. & 2.580 & 0.116 & 1.119 & 1.213 & 1.257 \\
        w/o Geo.      & 1.620 & 0.052 & 0.587 & 0.838 & 0.774 \\
        w/o Bounds    & 1.156 & 0.025 & 0.487 & 0.685 & 0.588 \\
        \textbf{Ours} & \textbf{1.121} & \textbf{0.021} & \textbf{0.431} & \textbf{0.619} & \textbf{0.565} \\
        \bottomrule
    \end{tabular}
    }
\end{table}

\begin{figure}[t]
    \centering
    \includegraphics[width=0.97\linewidth]{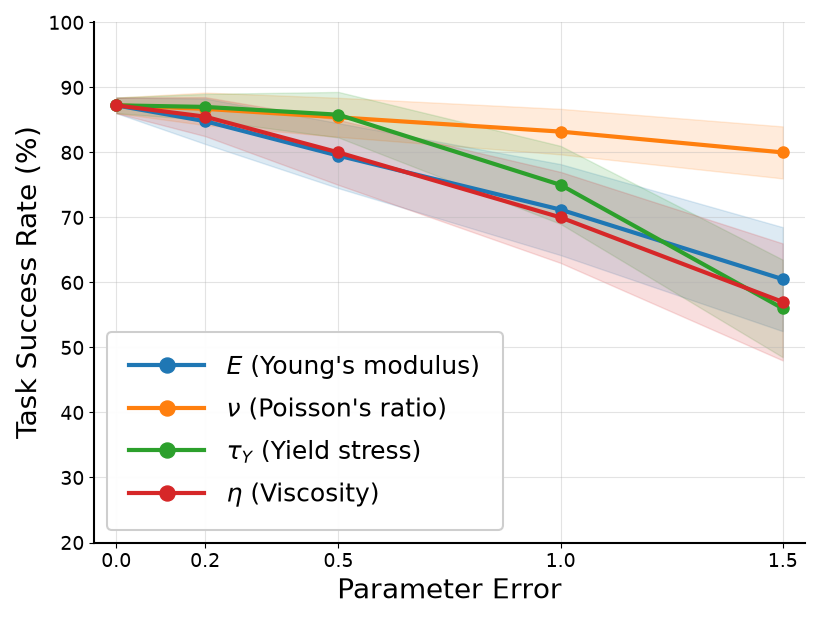}
    \caption{
Sensitivity analysis with respect to physical parameter estimation errors. Shaded regions indicate the uncertainty across repeated trials. Overall, the task success rate decreases with increasing estimation errors.
}
    \label{fig:parameter_sensitivity}
\end{figure}

\paragraph{Physical Parameter Estimation Accuracy.} We evaluate whether the proposed differentiable MPM framework can accurately recover the material parameters
$\boldsymbol{\theta}=\{E,\nu,\tau_Y,\eta\}$. For simulated objects, the estimated parameters are compared against the corresponding ground-truth values provided by the simulator. Following the large dynamic ranges of $E$, $\tau_Y$, and $\eta$, we report the mean absolute error in the logarithmic domain (log-MAE) for these parameters, while using the mean absolute error (MAE) for $\nu$. We compare the full model against four variants: \textit{(i) w/o VLM Init.}, which removes the VLM-based initialization; \textit{(ii) w/o Bounds}, which removes the material-aware lower and upper bounds during optimization; \textit{(iii) w/o Geo.}, which excludes the geometric consistency constraint $\mathcal{L}_{geo}$; and \textit{(iv) w/o BC}, which removes the manipulation boundary constraint $\mathcal{L}_{bc}$. 
As shown in Table~\ref{tab:physical_parameter_estimation}, our full model achieves the lowest estimation error across all physical parameters. Removing VLM initialization causes the largest degradation, demonstrating the importance of a physically meaningful initialization. Removing the material-aware bounds also reduces accuracy, as unconstrained optimization may enter implausible parameter regions. Similarly, excluding the geometric consistency constraint consistently increases the error, confirming that geometry-level supervision improves the alignment between simulated dynamics and observed deformations.

\paragraph{Sensitivity to Physical Parameter Errors.}
We investigate how errors in individual physical parameters affect
downstream manipulation performance. Starting from the ground-truth
parameters in simulation, we perturb one parameter at a time while
keeping all remaining parameters fixed. For $E$, $\tau_Y$, and $\eta$,
the perturbation is applied in the logarithmic domain as
$\log_{10}\tilde{\theta}_i =
\log_{10}\theta_i^{*}+\delta$, where
$|\delta|\in\{0,0.2,0.5,1.0,1.5\}$.
For each non-zero error level, we evaluate both positive and negative
perturbations and average their success rates. Since $\nu$ lies within
a narrow physical range, we perturb it additively and report its
normalized absolute error. The policy is kept frozen throughout the
evaluation, and only one parameter is modified in each experiment.
We report the average task success rate across all objects and tasks in Fig.~\ref{fig:parameter_sensitivity}. The success rate generally decreases as the estimation error increases, indicating that accurate physical parameter estimation is important for reliable downstream manipulation. The policy is relatively robust to errors in $\nu$, while errors in $E$, $\tau_Y$, and $\eta$ cause larger performance degradation, especially at high error levels.

\subsection{Robot Manipulation Benchmarks}
\label{sec:exp_manipulation}

\begin{table*}
    \centering
    \caption{
    Success rates (\%) of different methods on pick-and-place and stacking
    tasks in simulation and real-world environments.
    Results are grouped by four distinct material categories to ensure complete data coverage.
    }
    \label{tab:main_results}
    
    \resizebox{\textwidth}{!}{
    \begin{tabular}{lccccccccccccccc}
        \toprule
        
        \multirow{3}{*}{Method}
        
        & \multicolumn{7}{c}{Simulation}
        & \multicolumn{8}{c}{Real Robot} \\
        
        \cmidrule(lr){2-8}
        \cmidrule(lr){9-16}
        
        & \multicolumn{4}{c}{Pick-and-Place}
        & \multicolumn{3}{c}{Stacking}
        & \multicolumn{4}{c}{Pick-and-Place}
        & \multicolumn{4}{c}{Stacking} \\
        
        \cmidrule(lr){2-5}
        \cmidrule(lr){6-8}
        \cmidrule(lr){9-12}
        \cmidrule(lr){13-16}
        
        & Ela.
        & Plas.
        & Foam
        & Rig.
        & Plas.
        & Foam
        & Rig.
        
        & Ela.
        & Plas.
        & Foam
        & Rig.
        
        & Ela.
        & Plas.
        & Foam
        & Rig. \\
        
        \midrule
        
        DP
        & 52.1 & 97.5 & 89.2 & 94.2
        & 81.7 & 60.8 & 74.6
        & 25.2 & 28.0 & 28.3 & 34.2
        & 38.3 & 26.7 & 23.3  & 43.3 \\
        
        DP3
        & 62.5 & \textbf{98.3} & 93.3 & 96.7
        & 85.0 & 73.3 & 82.9
        & 30.0 & 43.3 & 46.7 & 49.2
        & 40.0 & 40.0 & 33.3  & 56.7 \\
        
        \textbf{Ours}
        & \textbf{70.8} & \textbf{98.3} & \textbf{95.0} & \textbf{97.2}
        & \textbf{87.5} & \textbf{82.5} & \textbf{85.8}
        & \textbf{57.1} & \textbf{56.0} & \textbf{66.7} & \textbf{70.0}
        & \textbf{56.7} & \textbf{60.0} & \textbf{46.7}  & \textbf{63.3} \\
        
        \bottomrule
    \end{tabular}
    }
    
    \vspace{1mm}
    
    \begin{minipage}{0.99\textwidth}
        \footnotesize
        \textit{Abbreviations:}
        Ela. = Elastic objects;
        Plas. = Plastic objects;
        Foam = Foam or sponge objects;
        Rig. = Rigid objects made of ceramic, metal, or wood.
    \end{minipage}
\end{table*}

\begin{table}[t]
\centering
\caption{
Success rates (\%) on objects with different material properties.
Ela. denotes elastic objects, Plas. denotes plastic objects,
Foam denotes foam or sponge objects, and Rig. denotes rigid objects
made of ceramic, metal, or wood.
}
\label{tab:object_success}
\resizebox{0.46\textwidth}{!}{
\begin{tabular}{llccc}
\toprule
Material
& Object
& DP
& DP3
& Ours \\
\midrule

\multirow{8}{*}{Ela.}
& Tennis Ball      & 33.3 & 43.3 & 60.0 \\
& Metal Spring     & 0.0  & 6.7  & 60.0 \\
& Shuttlecock      & 46.7 & 30.0 & 73.3 \\
& Stuffed Toy      & 6.7  & 13.3 & 33.3 \\
& Circular Bread   & 20.0 & 26.7 & 43.3 \\
& Elongated Bread  & 10.0 & 3.3  & 46.7 \\
& Tissue Package   & 60.0 & 86.7 & 83.3 \\
& Avg.             & 25.2 & 30.0 & \textbf{57.1} \\
\midrule

\multirow{6}{*}{Plas.}
& Banana           & 26.7 & 36.7 & 40.0 \\
& Grape            & 6.7  & 33.3 & 53.3 \\
& Toothpaste Tube  & 0.0  & 16.7 & 63.3 \\
& Disposable Cup   & 80.0 & 86.7 & 86.7 \\
& Ping-Pong Ball   & 26.7 & 43.3 & 36.7 \\
& Avg.             & 28.0 & 43.3 & \textbf{56.0} \\
\midrule

\multirow{2}{*}{Foam}
& Foam Ball        & 0.0  & 36.7 & 63.3 \\
& Washing Sponge        & 56.7  & 56.7 & 70.0 \\
& Avg.             & 28.0  & 46.7 & \textbf{66.7} \\
\midrule

\multirow{5}{*}{Rig.}
& Hard Candy         & 10.0 & 30.0 & 70.0 \\
& Tape Roll          & 6.7  & 23.3 & 33.3 \\
& Large Wooden Cube  & 56.7 & 73.3 & 90.0 \\
& Middle Wooden Cube & 63.3 & 70.0 & 86.7 \\
& Avg.               & 34.2 & 49.2 & \textbf{70.0} \\
\bottomrule
\end{tabular}
}
\end{table}

\paragraph{Multi-Material Performance.}
Table~\ref{tab:main_results} reports the success rates across four representative material categories in both simulation and real-world environments. Our method consistently achieves the highest success rates across both tasks, with particularly large improvements on deformable materials, where accurately modeling object dynamics is essential for stable manipulation. Notably, the performance gains are even more pronounced in the real-world experiments, where manipulation involves more complex object variations, sensing noise, and contact dynamics than in simulation. This demonstrates that our material-aware physical representations generalize effectively beyond controlled simulation settings and enable more robust manipulation in diverse real-world scenarios.

\begin{figure}[t]
    \centering
    \includegraphics[width=0.97\linewidth]{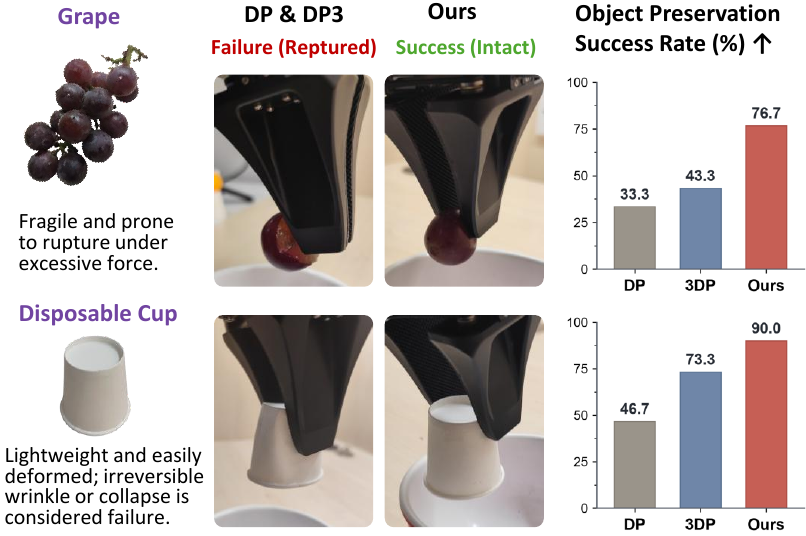}
    \caption{
Object preservation results on fragile objects. Among successful pick-and-place trials, the methods differ in their ability to preserve object integrity. We evaluate object preservation by measuring grape rupture and irreversible deformation of paper cups over 30 successful pick-and-place trials. A trial is considered successful only if the object remains intact, exhibiting neither grape rupture nor irreversible deformation. Our method achieves the highest object preservation success rate on both object categories.
}
    \label{fig:obj_preserve}
\end{figure}

\paragraph{Object Preservation on Fragile Objects.} Successful task completion does not necessarily imply successful manipulation of fragile objects. In real-world settings, objects such as grapes and paper cups may be damaged during grasping despite being successfully picked and placed. We therefore perform an additional evaluation on these two objects, where a trial is considered successful only if the object remains intact, i.e., without grape rupture or irreversible cup deformation. Each method is evaluated over 60 trials per object. As shown in Fig.~\ref{fig:obj_preserve}, our method substantially improves the object preservation success rate, indicating that the learned material-aware policy produces more compliant interactions with fragile objects while maintaining task success.

\subsection{Ablation Studies}
\label{sec:exp_ablations}

\begin{table*}[t]
    \centering
    \caption{
    Ablation study.
    We report task success rates (\%) across different material categories
    in simulation and real-world environments.
    ``Physics Cond.'' denotes conditioning the policy on the estimated
    physical parameters; ``Expert Pretrain.'' denotes independently
    pretraining the four material-specific diffusion experts before joint
    fine-tuning; and ``Adaptive Router'' denotes the learned physics-aware
    soft routing mechanism.
    Higher values indicate better performance.
    }
    \label{tab:moe_ablation}

    \resizebox{\textwidth}{!}{
    \begin{tabular}{lccc|ccccc|ccccc}
        \toprule

        \multirow{2}{*}{Variant}
        & \multirow{2}{*}{\makecell{Physics\\Cond.}}
        & \multirow{2}{*}{\makecell{Expert\\Pretrain.}}
        & \multirow{2}{*}{\makecell{Adaptive\\Router}}
        & \multicolumn{5}{c|}{Simulation}
        & \multicolumn{5}{c}{Real Robot} \\

        \cmidrule(lr){5-9}
        \cmidrule(lr){10-14}

        &
        &
        &
        &
        Ela. $\uparrow$
        & Plas. $\uparrow$
        & Foam $\uparrow$
        & Rig. $\uparrow$
        & Avg. $\uparrow$
        &
        Ela. $\uparrow$
        & Plas. $\uparrow$
        & Foam $\uparrow$
        & Rig. $\uparrow$
        & Avg. $\uparrow$ \\

        \midrule
    
        Single shared policy (DP)
        & \xmark
        & \xmark
        & \xmark
        & 52.1
        & 97.5
        & 89.2
        & 94.2
        & 83.3
        & 25.2 & 28.0 & 28.3 & 34.2
        & 28.9 \\

        Single shared policy + physics
        & \cmark
        & \xmark
        & \xmark
        & 69.2
        & 97.5
        & 93.3
        & 96.7
        & 87.9
        & 43.8
        & 44.7
        & 41.7
        & 67.5
        & 49.4 \\

        MoE w/o expert pretraining
        & \cmark
        & \xmark
        & \cmark
        & 60.8
        & 96.7
        & 93.3
        & 97.5
        & 85.8
        & 48.1
        & 34.7
        & 56.7
        & 66.7
        & 51.6 \\
    
        \textbf{Full model}
        & \cmark
        & \cmark
        & \cmark
        & \textbf{70.8}
        & \textbf{98.3}
        & \textbf{95.0}
        & \textbf{97.2}
        & \textbf{89.9}
        & \textbf{57.1}
        & \textbf{56.0}
        & \textbf{66.7}
        & \textbf{70.0}
        & \textbf{62.5} \\
   
        \bottomrule
    \end{tabular}
    }
\end{table*}

We conduct ablation studies to evaluate the contribution of each component. Specifically, we examine the effects of physics conditioning, expert pretraining, and the adaptive routing mechanism. Table~\ref{tab:moe_ablation} summarizes the results on both simulation and real-world pick-and-place tasks.

\paragraph{Effect of Physics Conditioning.}
We first investigate whether explicitly conditioning the diffusion policy on the estimated physical parameters benefits robot manipulation. Compared with the vanilla Diffusion Policy, introducing physics conditions consistently improves performance across almost all material categories, increasing the average success rate from 83.3\% to 87.9\% in simulation and from 28.9\% to 49.4\% on the real robot. The improvement is particularly pronounced for elastic and foam objects, whose manipulation strongly depends on material properties rather than appearance alone. These results demonstrate that the estimated physical parameters provide complementary information beyond visual observations and enable the policy to produce more appropriate interaction behaviors for objects with diverse physical characteristics.

\paragraph{Effect of Expert Pretraining.}
We further evaluate the importance of independently pretraining the material-specific diffusion experts before joint optimization. Removing expert pretraining consistently degrades performance, especially for elastic and plastic objects. Without dedicated initialization, the experts must simultaneously learn both material-specific manipulation behaviors and expert specialization during joint training, leading to unstable optimization and reduced generalization. In contrast, pretraining each expert on demonstrations from its corresponding material category provides a strong initialization, allowing subsequent joint optimization to focus on learning collaboration among experts and improving overall policy performance.

\paragraph{Effect of Adaptive Material Routing.}
Finally, we evaluate the proposed physics-aware routing mechanism. Although conditioning a single shared policy on physical parameters already improves over the vanilla Diffusion Policy, it still requires one network to model diverse deformation and contact dynamics across different materials. As a result, the shared policy tends to overfit to dominant interaction patterns, achieving strong performance on familiar objects while generalizing poorly to objects with distinct physical behaviors. In contrast, the proposed adaptive router dynamically assigns material-specific experts according to the estimated physical properties, enabling more specialized policy learning.

\begin{figure}[t]
    \centering
    \includegraphics[width=\linewidth]{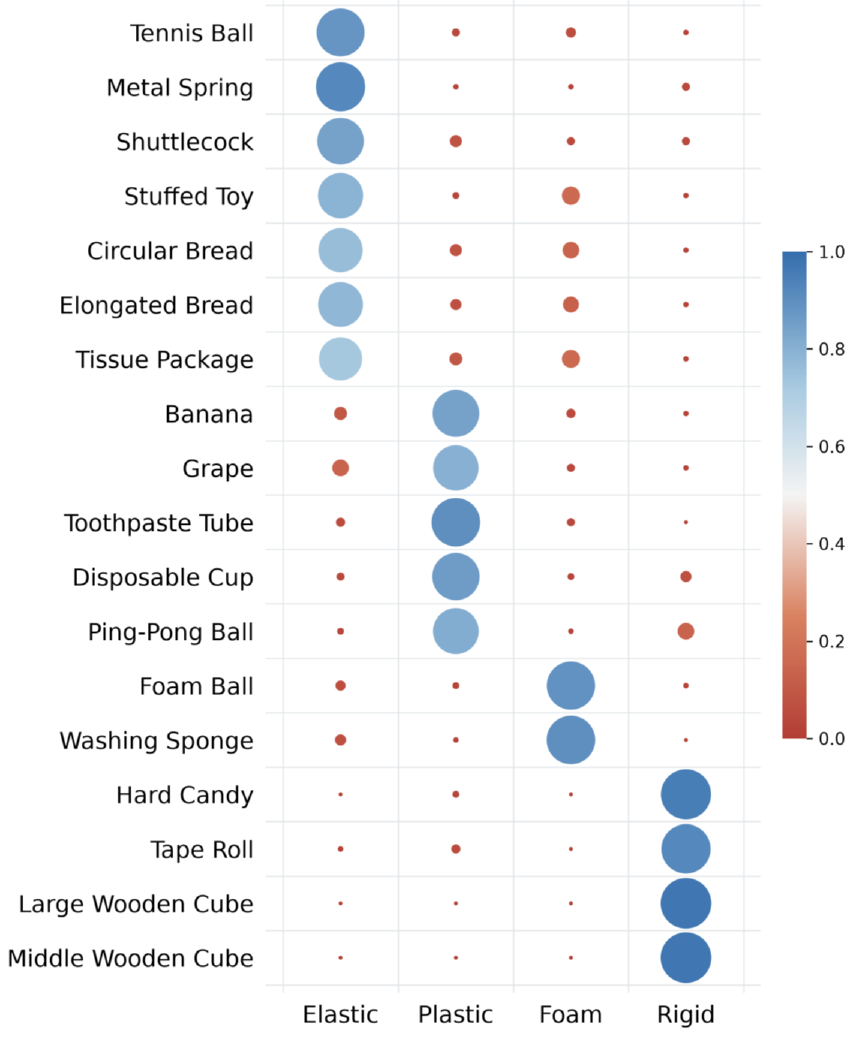}
    \caption{
    Visualization of learned material-aware routing weights in our MoE policy.
    Each row represents an object and each column corresponds to a physical-property expert.
    The bubble size and color indicate the routing weight assigned to each expert.
    The learned router discovers clear expert specialization, where objects with different physical characteristics are preferentially routed to their corresponding experts while maintaining adaptive soft assignments across experts.
    }
    \label{fig:moe_routing}
\end{figure}

\paragraph{Expert Routing Visualization.}
To better understand the learned routing behavior, we visualize the routing weights produced by the physics-aware router in Fig.~\ref{fig:moe_routing}. 
Different object categories consistently activate different experts while maintaining soft assignments across multiple experts, indicating that the router learns meaningful material-aware specialization. Such adaptive routing enables smooth transitions between different physical regimes and contributes to the strong generalization performance.

\paragraph{MoE Policy Architecture.}
We ablate the proposed material-aware MoE policy to evaluate the effect of expert specialization and adaptive routing. We compare our full model against a single-expert policy. As shown in Table~\ref{tab:moe_ablation}, removing expert specialization degrades performance across diverse materials. The shared policy tends to overfit to a subset of objects, achieving high accuracy on familiar interaction patterns but significantly worse performance on objects with different physical dynamics. In contrast, the learned router enables material-aware expert selection, improving adaptation to heterogeneous objects.

\subsection{Memory Bank and Online Inference Efficiency}
\label{sec:exp_memory_bank}

\begin{table}[t]
\centering
\caption{
Retrieval performance under the within-domain evaluation setting across
different view configurations.
All results are reported as percentages.
Higher values indicate better retrieval performance.
}
\label{tab:retrieval_performance}
\resizebox{0.48\textwidth}{!}{
\begin{tabular}{lcccccc}
\toprule
\multirow{2}{*}{Configuration}
& \multicolumn{2}{c}{Instance Retrieval}
& \multicolumn{2}{c}{Material Retrieval}
& \multicolumn{2}{c}{Physics Retrieval} \\
\cmidrule(lr){2-3}
\cmidrule(lr){4-5}
\cmidrule(lr){6-7}
& Hit@1 $\uparrow$
& Hit@3 $\uparrow$
& Hit@1 $\uparrow$
& Hit@3 $\uparrow$
& Recall@1 $\uparrow$
& Recall@3 $\uparrow$ \\
\midrule

Static only
& 93.2 & 99.0
& 97.1 & 99.6
& 94.8 & 99.2 \\

Wrist only
& 68.7 & 84.3
& 79.4 & 91.6
& 74.8 & 88.5 \\

\textbf{Static + Wrist}
& \textbf{96.8} & \textbf{99.7}
& \textbf{98.8} & \textbf{100.0}
& \textbf{97.4} & \textbf{99.8} \\

\bottomrule
\end{tabular}
}
\end{table}

\begin{table}[t]
    \centering
    \caption{
    Comparison of online inference efficiency and downstream manipulation
    performance between online physics optimization and memory-bank retrieval. We report the average time per query.
    Success rates (\%) are reported for pick-and-place and stacking
    tasks in real-world environments.
    }
    \label{tab:online_inference_efficiency}
    \resizebox{0.48\textwidth}{!}{
    \begin{tabular}{lccccc}
        \toprule
        Method
        & Bank Size
        & Time
        & Pick-and-Place
        & Stacking
        & Avg.
        \\
        \midrule

        Optimization
        & N/A
        & 798.22 s
        & 63.3
        & 58.0
        & 60.7 \\

        \midrule

        Bank Retrieval
        & 15
        & 414.90 ms
        & 61.6
        & 56.7
        & 59.2
         \\

           Bank Retrieval
        & 50
        & 414.91 ms
        & 61.6
        & 56.0
        & 58.8
         \\

           Bank Retrieval
        & 200
        & 414.93 ms
        & 60.0
        & 54.7
        & 57.4
        \\

        \bottomrule
    \end{tabular}
    }
\end{table}


\paragraph{Retrieval Accuracy.}
We evaluate retrieval quality at the instance, material, and physics levels.
Instance Hit@$K$ measures whether the exact query object appears among the
top-$K$ retrieved entries, while Material Hit@$K$ checks whether at least one
retrieved entry belongs to the same material category. Physics Recall@$K$
measures whether the top-$K$ results contain an entry whose normalized physical
parameter distance is below $\epsilon=1.0$, where each parameter is normalized
by the estimation error of our full model in
Table~\ref{tab:physical_parameter_estimation}. All metrics are averaged over
the test queries.

As shown in Table~\ref{tab:retrieval_performance}, the static view consistently
outperforms the wrist view, as it provides more stable global observations and
is less affected by occlusion and viewpoint variation. Combining the two views
achieves the best performance across all metrics, demonstrating that they
provide complementary visual cues. 
In particular, its strong Hit@1 and
Physics Recall@1 performance motivates the direct use of the top-1 retrieved entry
for policy conditioning.

\paragraph{Latency and Computational Overhead.}
As shown in Table~\ref{tab:online_inference_efficiency}, online physics optimization requires 798.22 s per query, making it unsuitable for real-time deployment. In contrast, memory-bank retrieval takes only $\sim$415 ms, achieving nearly 1900$\times$ faster inference while maintaining almost constant latency across different bank sizes. Despite eliminating online optimization, retrieval preserves comparable manipulation performance. With a bank size of 15, the average success rate reaches 59.2\%, only 1.5\% lower than online optimization (60.7\%). These results demonstrate that offline memory-bank retrieval effectively approximates online physics optimization, enabling real-time deployment with minimal performance degradation.

%% file: sections/5_con.tex
\section{Conclusion}

We presented a physics-aware diffusion policy that bridges differentiable physical simulation and diffusion-based robot manipulation. Our framework estimates object-specific material properties from visual demonstrations through a differentiable 3DGS-MPM representation, conditions a material-aware Mixture-of-Experts diffusion policy on the recovered physical parameters, and enables efficient deployment via a physics memory bank. Extensive experiments in both simulation and real-world environments demonstrate improved manipulation success, object preservation, and robustness across diverse objects with heterogeneous physical properties. The results suggest that explicitly modeling object physics significantly enhances the scalability and generalization of diffusion policies beyond purely appearance-based imitation learning. More broadly, we believe that integrating differentiable physical reasoning with generative policy learning is a promising direction toward general-purpose robot manipulation in complex real-world environments.